\documentclass{article}
\usepackage{amsmath}

\usepackage{arxiv}

\usepackage[utf8]{inputenc} 
\usepackage[T1]{fontenc}    
\usepackage{booktabs}       
\usepackage{amsfonts}       
\usepackage{nicefrac}       
\usepackage{graphicx}
\usepackage{doi}

\usepackage{booktabs}
\usepackage{pifont}
\usepackage[table]{xcolor} 
\usepackage{threeparttable}

\title{OphthBench: A Comprehensive Benchmark for Evaluating Large Language Models in Chinese Ophthalmology}

\author{ Chengfeng~Zhou \\
	Changsha Aier Eye Hospital\\
	Changsha, China \\
	\texttt{joe1chief1993@gmail.com} \\
	\And
	Ji~Wang \\
	Changsha Aier Eye Hospital\\
	Changsha, China \\
	\texttt{wangji\_oph@163.com} \\
	\And
        Juanjuan~Qin \\
	Changsha Aier Eye Hospital\\
	Changsha, China \\
	\texttt{769773911@qq.com} \\
	\And
        Yining~Wang \\
	Changsha Aier Eye Hospital\\
	Changsha, China \\
	\texttt{wangyining2@aierchina.com} \\
	\And
        Ling~Sun \\
	Changsha Aier Eye Hospital\\
	Changsha, China \\
	\texttt{2874351123@qq.com} \\
	\And
        Weiwei~Dai \\
	Changsha Aier Eye Hospital\\
	Changsha, China \\
	\texttt{daiweiwei@aierchina.com}\\
}

\renewcommand{\shorttitle}{\textit{arXiv} Template}

\hypersetup{
pdftitle={A template for the arxiv style},
pdfsubject={q-bio.NC, q-bio.QM},
pdfauthor={David S.~Hippocampus, Elias D.~Striatum},
pdfkeywords={First keyword, Second keyword, More},
}

\begin{document}
\maketitle

\begin{abstract}
Large language models (LLMs) have shown significant promise across various medical applications, with ophthalmology being a notable area of focus.
Many ophthalmic tasks have shown substantial improvement through the integration of LLMs.
However, before these models can be widely adopted in clinical practice, evaluating their capabilities and identifying their limitations is crucial. 
To address this research gap and support the real-world application of LLMs, we introduce the OphthBench, a specialized benchmark designed to assess LLM performance within the context of Chinese ophthalmic practices. 
This benchmark systematically divides a typical ophthalmic clinical workflow into five key scenarios: Education, Triage, Diagnosis, Treatment, and Prognosis.
For each scenario, we developed multiple tasks featuring diverse question types, resulting in a comprehensive benchmark comprising 9 tasks and 591 questions.
This comprehensive framework allows for a thorough assessment of LLMs' capabilities and provides insights into their practical application in Chinese ophthalmology.
Using this benchmark, we conducted extensive experiments and analyzed the results from 39 popular LLMs.
Our evaluation highlights the current gap between LLM development and its practical utility in clinical settings, providing a clear direction for future advancements. 
By bridging this gap, we aim to unlock the potential of LLMs and advance their development in ophthalmology.
\end{abstract}


\section{Introduction}
\label{sec:introduction}
Large language models (LLMs) have revolutionized a wide range of applications driven by their remarkable ability to understand and interpret real-world contexts~\cite{brown2020language,achiam2023gpt,singhal2023large,gao2024empowering,chen2024survey}.
In the medical domain, LLMs have demonstrated significant potential across diverse applications, such as disease management~\cite{li2024integrated}, medical document summarization~\cite{tang2023evaluating,van2024adapted}, clinical workflows optimization~\cite{klang2024strategy}, and patient-clinical trial matching~\cite{jin2024matching,pmlr-v219-wong23a}.
Despite these advancements, ongoing debates persist regarding the ethical use of LLM technology in healthcare, addressing concerns such as ``hallucination''(generating plausible-sounding but inaccurate or misleading content), biases inherent in training data, and broader ethical considerations~\cite{shen2023chatgpt,li2023ethics,haltaufderheide2024ethics,luo2024clinical}.
Accurately assessing LLMs' potential and inherent limitations is urgently needed to ensure their beneficial use for patients and practitioners.

Current research that evaluates LLMs against specific benchmarks primarily focuses on their general capabilities, as demonstrated by tasks such as medical question answering~\cite{pal2022medmcqa}, medical calculations~\cite{khandekar2024medcalc}, comprehension of lengthy clinical documents~\cite{adams2024longhealth}, and named entity recognition~\cite{zhang2021cblue}.
While these approaches help guide the evolution of LLM algorithms, it remains unclear whether their broad capabilities translate into practical medical utility.
Consequently, validating the feasibility of LLMs in real-world clinical scenarios is an urgent priority to ensure their more effective application.

Numerous researchers have examined the comprehensive capabilities of LLMs in specific clinical tasks ~\cite{suthar2023artificial,goodman2023accuracy,antaki2023capabilities,bernstein2023comparison,hager2024evaluation} and disease contexts ~\cite{rao2023assessing,lim2023benchmarking,li2023chatgpt,longwell2024performance,huang2024assessment}, often estimating their performance through carefully designed question sets merely.
However, such efforts alone are insufficient for achieving a standardized and holistic evaluation, indicating the need for a robust and widely accepted evaluation benchmark.
Here, a benchmark is defined as a systematic and standardized set of tasks and evaluation protocols against which the performance of LLMs can be consistently assessed and compared.
Developing such a benchmark is by no means trivial.
First, different regions exhibit variations in cultural backgrounds, legal frameworks, and healthcare systems that shape distinct medical practices—factors that must be accounted for~\cite{DBLP:journals/corr/abs-2009-13081,zhang2021cblue,DBLP:conf/chil/PalUS22,liu2024benchmarking,DBLP:conf/aaai/0001WWMZWH24,jiang2024jmedbench}.
Second, current benchmarks designed to evaluate general medical capabilities often lack the depth required to reflect the realities of clinical practice, emphasizing the need for specialized benchmarks that capture the complexities inherent in specific applications~\cite{DBLP:conf/kdd/ChenMGWZC24,DBLP:journals/corr/abs-2406-01126,liu2024benchmarking}.
Third, the inherent issues of LLMs, prompt sensitivity~\cite{zhuo2024prosa} and data contamination~\cite{dong2024generalization}, hamper the consistent, objective evaluations, for which a carefully designed protocol is needed to ensure fair comparisons.

To address the aforementioned challenges, we introduce OphthBench, a specialized benchmark for assessing LLM capabilities in Chinese ophthalmology. 
Developed by three experienced Chinese ophthalmologists, it comprises $591$ questions spanning $5$ core ophthalmic scenarios and evaluates model performance across $9$ distinct tasks.
The integrated expertise in the benchmark ensures alignment with real-world clinical practice in China and provides a valid measure for the practical clinical utility of LLMs.
Using this benchmark, we evaluate $39$ LLMs, encompassing general-purpose or medical-specific, monolingual or multilingual, and open-source or commercial models. 
We conduct a comprehensive analysis that clarifies their clinical utility in Chinese ophthalmology.
In summary, the main contributions of this paper are as follows:
\begin{enumerate}
    \item We introduce a well-designed benchmark for assessing the clinical utility of LLMs in Chinese Ophthalmology, which comprises $5$ key scenarios and $9$ distinct tasks.
    \item We present diverse problem settings and a rigorous evaluation protocol that ensures comprehensive assessment and mitigates the influence of prompt sensitivity.
    \item We evaluate $39$ popular LLMs using our benchmark, revealing the substantial gap between current model capabilities and practical requirements, as well as pointing out their potential evolution directions.
\end{enumerate}

\begin{figure*}[!t]
    \centering
    \includegraphics[width=1.0\textwidth]{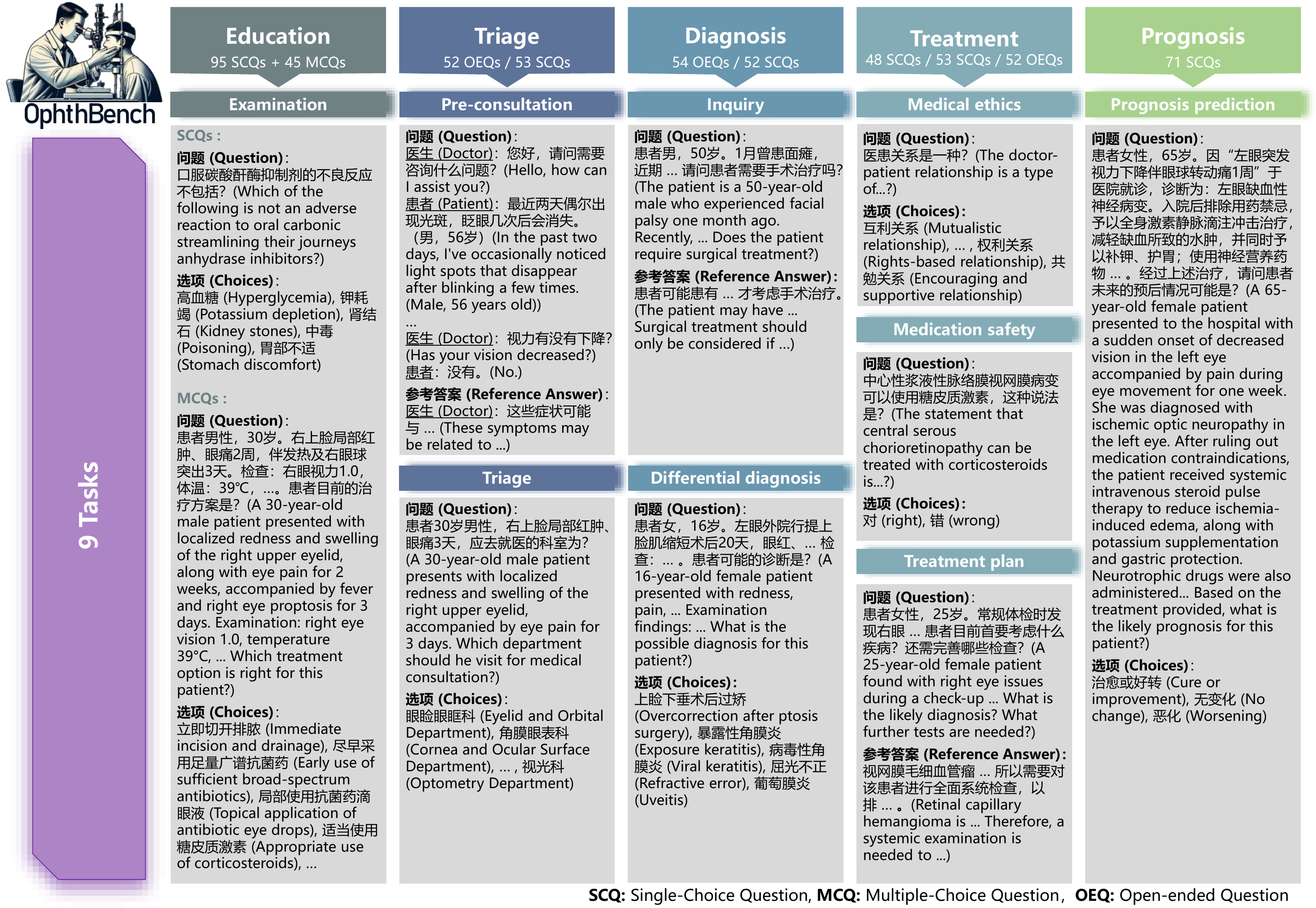}
    \caption{\textbf{An illustration of our proposed benchmark.} OphthBench is a specialized benchmark designed to assess LLM capabilities in Chinese ophthalmic practices. It comprises $5$ core ophthalmic scenarios, \textbf{Education}, \textbf{Triage}, \textbf{Diagnosis}, \textbf{Treatment}, and \textbf{Prognosis}, covering $9$ distinct tasks with single-choice, multiple-choice, and open-ended question formats. This comprehensive structure ensures a thorough evaluation of LLM performance across various key areas in ophthalmic practice.}
\label{fig:overview}
\end{figure*}

\section{Related Work}
Benchmarking is a critical step in advancing large language models, serving as a cornerstone to guide their application and development. However, as expectations for LLMs continue to rise, general language understanding benchmarks no longer suffice for specialized professional domains.
In the medical domain, numerous datasets and benchmarks~\cite{pal2022medmcqa,zhang2021cblue,suthar2023artificial,singhal2023large,zhu2023promptcblue,goodman2023accuracy,rao2023assessing,lim2023benchmarking,li2023chatgpt,wang2023cmb,liu2024benchmarking,hager2024evaluation,longwell2024performance,DBLP:conf/aaai/0001WWMZWH24} have emerged to evaluate the medical capabilities of LLMs. 
These studies aim to enhance clinical utility and alleviate the workload of healthcare professionals. 
For example, Pal et al.~\cite{pal2022medmcqa} established their benchmark with $194$k high-quality AIIMS \& NEET PG entrance exam multiple-choice questions, and Wang et al.~\cite{wang2023cmb} introduced the CMB, a localized medical benchmark designed entirely within the native Chinese linguistic and cultural framework.

Among these studies~\cite{pal2022medmcqa,zhang2021cblue,suthar2023artificial,singhal2023large,zhu2023promptcblue,goodman2023accuracy,rao2023assessing,lim2023benchmarking,li2023chatgpt,wang2023cmb,liu2024benchmarking,hager2024evaluation,longwell2024performance,DBLP:conf/aaai/0001WWMZWH24}, multiple-choice questions are typically preferred because they simplify the evaluation process and offer an objective assessment.
In contrast, open-ended questions require subjective input from medical experts, which can be time-consuming and labor-intensive, yet they provide greater flexibility and deeper insights into an LLM's capabilities.
With the development of ``LLM-as-a-judge'' methods~\cite{son2024llm,cao2024compassjudger}, the dependence on human evaluation has been reduced, allowing assessments to be conducted automatically by LLMs.
Therefore, combining multiple-choice and open-ended questions within a single benchmark is preferable to provide a more comprehensive assessment of an LLM's capabilities~\cite{li2024multiple}. 

Despite the above progress, evaluating LLMs in the professional medical field remains challenging due to variations in specialized knowledge across different departments or diseases, as well as regional differences in medical systems and language use.
To address these complexities, benchmarks have become increasingly specialized, focusing on particular departments~\cite{zhang2024pediabench,DBLP:journals/corr/abs-2406-01126}, regions~\cite{DBLP:conf/aaai/0001WWMZWH24,jiang2024jmedbench}, or diseases~\cite{DBLP:conf/kdd/ChenMGWZC24}.
With regard to departmental disparities, Zhang et al.~\cite{zhang2024pediabench} introduced a Chinese pediatric dataset to compensate for the limited evaluations of LLMs in pediatrics.
Focusing on disease types, Chen et al.~\cite{DBLP:conf/kdd/ChenMGWZC24} examined rare diseases to investigate the capabilities and limitations of LLMs.
Meanwhile, concerning region-specific contexts,  Cai et al.~\cite{DBLP:conf/aaai/0001WWMZWH24} proposed MedBench to accommodate China's linguistic, and clinical standards and procedures. 

Although several studies have investigated LLMs' practical utility in ophthalmology~\cite{lim2023benchmarking,antaki2023capabilities,bernstein2023comparison,huang2024assessment}, they often lack comprehensiveness and a standardized evaluation framework and alignment with the ophthalmic clinical workflow.
There is an urgent need for an objective, systematic, and standardized benchmark tailored to the unique characteristics of Chinese ophthalmology to accurately assess LLM performance and its clinical applicability. 
To address this challenge, we propose OphthBench, a specialized benchmark incorporating diverse question types, designed to comprehensively assess LLM capabilities in this context, as summarized in Table~\ref{tab:benchmarks}.

\begin{table}[tb!]
    \centering
    \caption{OphthBench comprehensively incorporates the core clinical workflow and various question types.}
    \begin{tabular}{lcccc}
        \toprule
        \textbf{Benchmark} & \textbf{SCQ}  & \textbf{MCQ} & \textbf{QA} & \textbf{Clinical Workflow} \\
        \midrule
        CMExam~\cite{liu2024benchmarking} & \textcolor{green}{\ding{51}}& \textcolor{red}{\ding{55}}& \textcolor{green}{\ding{51}} & \textcolor{red}{\ding{55}}\\
        CMB~\cite{wang2023cmb} & \textcolor{green}{\ding{51}}& \textcolor{red}{\ding{55}}& \textcolor{green}{\ding{51}} & \textcolor{red}{\ding{55}}\\
        MedMCQA~\cite{pal2022medmcqa} & \textcolor{green}{\ding{51}}& \textcolor{red}{\ding{55}}& \textcolor{green}{\ding{51}} & \textcolor{red}{\ding{55}}\\
        MedBench (paper)~\cite{DBLP:conf/aaai/0001WWMZWH24} & \textcolor{green}{\ding{51}}& \textcolor{green}{\ding{51}}& \textcolor{red}{\ding{55}} & \textcolor{red}{\ding{55}}\\
        MedBench (website) & \textcolor{green}{\ding{51}}& \textcolor{green}{\ding{51}}& \textcolor{green}{\ding{51}} & \textcolor{red}{\ding{55}}\\
        MultiMedQA~\cite{singhal2023large} & \textcolor{green}{\ding{51}}& \textcolor{red}{\ding{55}}& \textcolor{green}{\ding{51}}& \textcolor{red}{\ding{55}}\\
        PromptCBLUE~\cite{zhu2023promptcblue} & \textcolor{green}{\ding{51}}& \textcolor{green}{\ding{51}}& \textcolor{green}{\ding{51}} & \textcolor{red}{\ding{55}}\\
        OphthBench (Ours) & \textcolor{green}{\ding{51}} & \textcolor{green}{\ding{51}}& \textcolor{green}{\ding{51}} & \textcolor{green}{\ding{51}}\\
        \bottomrule
    \end{tabular}
    \label{tab:benchmarks}
\end{table}

\section{The Proposed Benchmark}
LLMs have garnered significant attention within the medical community, primarily due to their expert-level performance on standardized medical question-answering tasks, including licensing exams \cite{liu2024benchmarking}, case studies \cite{haltaufderheide2024ethics}, and diseases diagnoses~\cite{savage2024diagnostic}.
In the field of ophthalmology, Betzler et al.~\cite{betzler2023large} highlighted the potential of LLMs for clinical practice and ophthalmic education. 
However, due to the lack of qualified evaluation benchmarks, the practical utility of LLMs in real-world Chinese ophthalmology scenarios cannot be quantified. 
To this end, we present a specialized benchmark for Chinese ophthalmology with the effort of three experienced ophthalmologists.

\subsection{The Taxonomy of OphthBench}
Drawing on the current clinical workflow in tertiary eye centers \cite{betzler2023large} and as illustrated in Fig.~\ref{fig:overview}, we identify five key application scenarios for LLMs in ophthalmology: education, triage, diagnosis, treatment, and prognosis.

\textbf{Education:} This scenario involves a broad range of activities, such as patient education, continuing medical education, and academic instruction for medical students and residents. 
To streamline the design, we dedicate our efforts exclusively to the \textit{examination} task, as it is essential for disseminating knowledge and effectively assesses the LLM's ability to retain ophthalmic information that underpins other activities.

\textbf{Triage:} This scenario focuses on assessing the urgency of a patient's condition to determine the priority of care. Based on this principle, two main tasks are defined: a \textit{pre-consultation} task, which assists patients in conducting an initial self-assessment, and a \textit{triage} task, which directs patients to the appropriate department according to their symptoms.

\textbf{Diagnosis:} This scenario focuses on identifying ocular diseases or conditions based on clinical data. 
The LLM's ability to complement an ophthalmologist's expertise is evaluated through an \textit{inquiry} task, which provides additional information to support clinical decisions. 
In parallel, a \textit{differential diagnosis} task assesses the automated diagnosis capability to distinguish among potential ocular conditions.

\textbf{Treatment:} This scenario emphasizes summarizing treatment guidelines and recommending appropriate medication regimens or other treatments based on the patient's clinical data. 
Accordingly, beyond the \textit{treatment plan} task, we also include two additional tasks: \textit{medical ethics} and \textit{medication safety} to evaluate the overall safety and reliability of the LLMs.

\textbf{Prognosis:} This scenario focuses on predicting disease progression to enhance risk management.
Therefore, we introduce a \textit{prognosis prediction} task, which asks LLMs to estimate patient outcomes following treatment.

Finally, these five scenarios collectively enable a more fine-grained evaluation of LLMs in ophthalmic practice and guide their future development.

\subsection{Construction and Statistics}
OphthBench is developed using representative exercises from the Chinese Medical Licensing Exam (CNMLE), Resident Standardization Training Exam, Doctor in-charge Qualification Exam, authoritative Chinese ophthalmology textbooks, and real-world clinical cases, ensuring that they accurately reflect Chinese ophthalmic practice. 
Three experienced ophthalmologists participated in the construction process: three junior ophthalmologists created and refined questions based on the aforementioned materials, while a senior ophthalmologist with an associate-level title reviewed all questions to ensure accuracy and quality.

\begin{figure}[!tb]
    \centering
    \includegraphics[width=\linewidth]{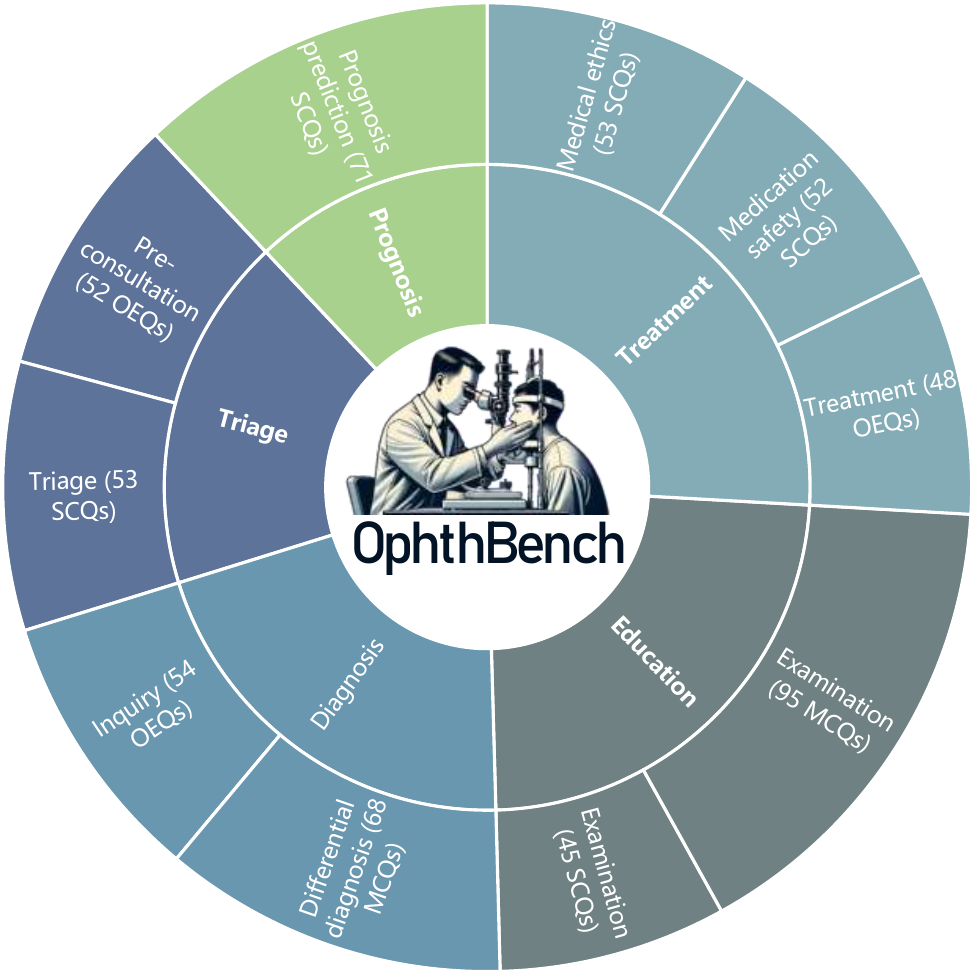}
    \caption{\textbf{The statistics of the OphthBench dataset.} With the help of three experienced Chinese ophthalmologists, OphthBench comprises $591$ questions spanning $5$ core ophthalmic scenarios and evaluates model performance across $9$ distinct tasks.} 
\label{fig:statistics}
\end{figure}

OphthBench includes three question types: single-choice questions (SCQs), multiple-choice questions (MCQs), and open-ended questions (OEQs). 
Answers to SCQs and MCQs are provided by ophthalmologists with their agreement, guaranteeing reliability and clinical relevance. 
For OEQs, initial answers are generated by GPT-4o, and then thoroughly reviewed and refined by ophthalmologists to maintain accuracy and align with the style of typical LLM responses.
Each application scenario in OphthBench is associated with an adequate number of questions for a comprehensive and satisfactory evaluation.
Fig.~\ref{fig:statistics} presents an overview of the dataset's statistics.

\begin{figure*}[!tb]
    \centering
    \includegraphics[width=1.0\textwidth]{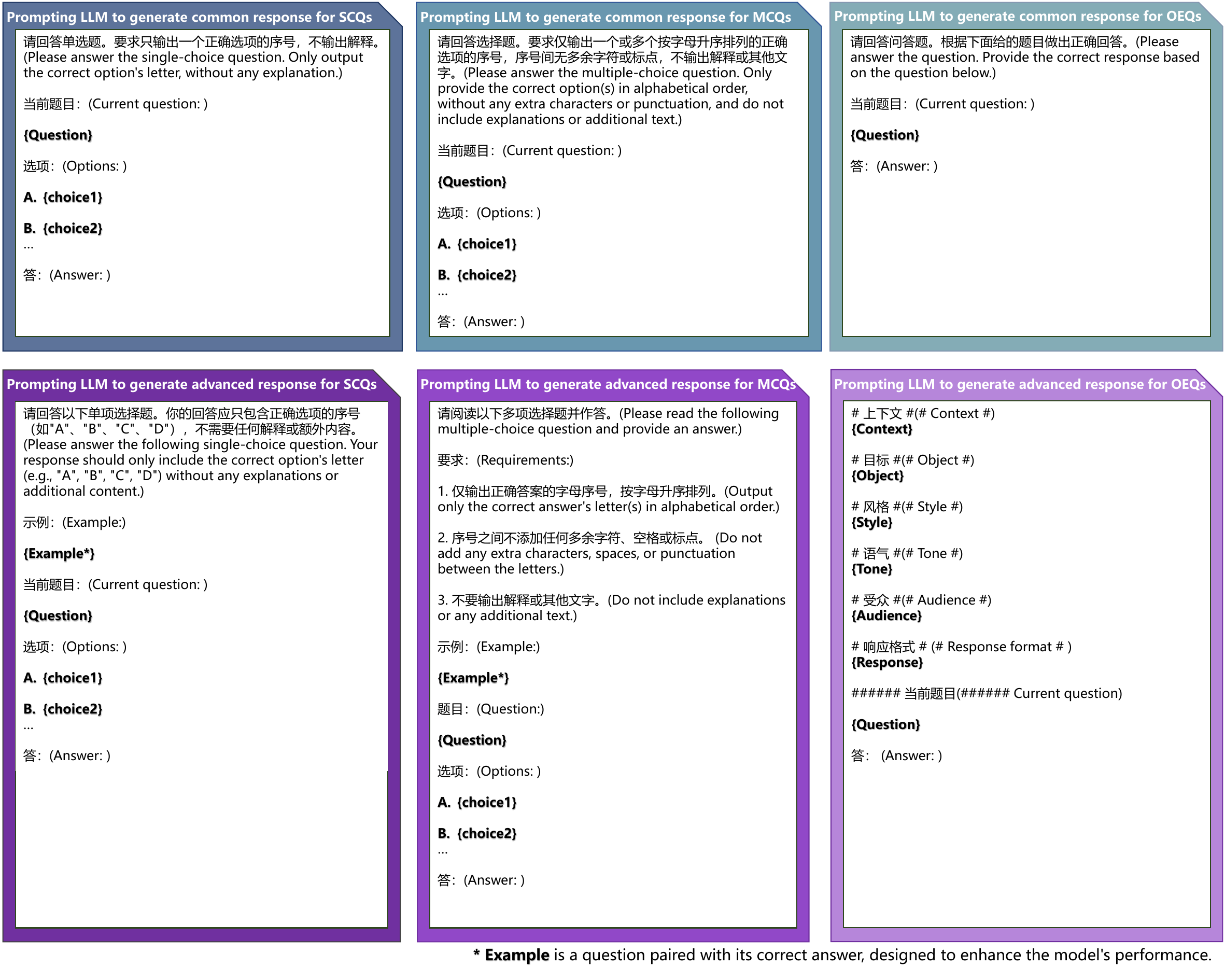}
    \caption{\textbf{The prompts utilized for evaluation.} Bold fonts indicate content that needs to be filled in. English translations in parentheses are not included during the evaluation.}
\label{fig:prompt}
\end{figure*}

\subsection{Evaluation Protocol}
Numerous studies have highlighted that the sensitivity of large language models to prompts can lead to biased evaluations. 
To mitigate these effects, we report two distinct performances for LLMs. 
The first uses a straightforward, concise prompt to mimic typical LLM common usage. 
The second applies advanced prompting techniques to optimize LLM performance.
As shown in Fig.~\ref{fig:prompt}, we use standard instructions to guide LLMs in generating common responses across various question types.
Then, we refine the above prompts with GPT4-o and add a fixed example to enhance performance for SCQs and MCQs.
For OEQs, the prompts for advanced LLM performance are tailored to the tasks, which follow the CO-STAR framework~\cite{teo2023won} and are auto-generated by GPT4-o using three randomly selected QA pairs.

To evaluate the LLMs' performance, we employ \textit{Accuracy} (\textit{Acc.}) for SCQs and $F_{1} $ $score$ for MCQs. 
Given that evaluating OEQs presents greater challenges, we adopt a rule-calibrated multi-dimensional point-wise LLM-as-Judge method~\cite{DBLP:conf/acl/LiuLWHFWCKXTZ0G24}, which automatically assesses the responses of LLMs in a human-like manner.
Specifically, \texttt{CompassJudger-1-7B}~\cite{cao2024compassjudger} assesses each response by comparing it to its reference answer across seven dimensions: Factuality, User Satisfaction, Safety and Harmlessness, Clarity, Responsibility, Logical Coherence, and Richness.

Before aggregating task performance, the highest score among $39$ LLMs for each task (on a scale of $100$) is set to $90$ and used for score normalization to eliminate the task difficulty bias.
Finally, the total and scenario scores for LLMs are calculated as the harmonic mean across their respective tasks.
This procedure can be formatted as follows:
\begin{equation}
    \begin{aligned}
    Score^{'}_{task} &= Score_{task} / Score^{\star}_{task} \times 90, \\
    Score^{'}_{scenario} &= \frac{n}{\sum_{task_{i} \in scenario}\frac{1}{Score^{'}_{task_{i}}}}, \\
    \end{aligned}
\end{equation}
, where $n$ is the number of tasks within a specific clinical $scenario$ and $Score^{\star}_{task}$ denotes the highest score among the $39$ LLMs with advanced reponses for the given $task$.

\section{Main Results}
A total of $39$ recent LLMs, including both open-sourced and commercial, were selected to evaluate their clinical utility in Chinese ophthalmology. 
These LLMs are the most widely used, and their basic information is summarized in Table~\ref{tab:model_description}.

We accessed commercial LLMs through their APIs, while for open-source models, we relied on the \texttt{transformers}\footnote{https://github.com/huggingface/transformers} library for local deployment.
We adopted regular expressions to extract option numbers from the commercial LLM responses to ensure each model could respond in a multiple-choice format. For open-source models, the \texttt{Outlines}\footnote{https://github.com/dottxt-ai/outlines} library was used to constrain outputs to specific options, thereby guaranteeing consistent answer formats.

\begin{table}[htb]
    \centering 
    \setlength{\tabcolsep}{1pt}
    \caption{LLMs evaluated with OphthBench.}
    \begin{tabular}{lcccc}
        \toprule
        \textbf{Model} & \textbf{Size} & \textbf{Version} & \textbf{Made in China} & \textbf{Medical use} \\
        \midrule
        \multicolumn{5}{c}{Open-source LLMs (on-device deployment)} \\
        \midrule
        Baichuan2-7B~\cite{yang2023baichuan}& 7B & Chat  & \textcolor{green}{\ding{51}}&  \textcolor{green}{\ding{55}}\\
        HuatuoGPT2-7B~\cite{zhang2023huatuogpt} & 7B & chat & \textcolor{green}{\ding{51}}& \textcolor{green}{\ding{51}}\\
        HuatuoGPT-o1-7B~\cite{chen2024huatuogpt} & 7B & - & \textcolor{green}{\ding{51}}& \textcolor{green}{\ding{51}}\\ 
        Llama-3.1-Tulu-3-8B~\cite{lambert2024t} & 8B &  - & \textcolor{red}{\ding{55}}& \textcolor{green}{\ding{55}}\\
        Meta-Llama-3.1-8B~\cite{dubey2024llama} & 8B & 8B-Instruct & \textcolor{red}{\ding{55}}& \textcolor{red}{\ding{55}}\\
        Ministral-8B~\cite{jiang2023mistral} & 8B & Instruct-2410 & \textcolor{red}{\ding{55}}& \textcolor{green}{\ding{55}}\\
        Mistral-7B~\cite{jiang2023mistral} & 7B & Instruct-v0.3 &  \textcolor{red}{\ding{55}}& \textcolor{red}{\ding{55}}\\
        PULSE-7B~\cite{pulse2023} & 7B & 7bv5 & \textcolor{green}{\ding{51}}& \textcolor{green}{\ding{51}}\\
        Phi-3.5-mini~\cite{abdin2024phi} & 3.82B & instruct & \textcolor{red}{\ding{55}}& \textcolor{red}{\ding{55}}\\
        Qwen2.5-7B~\cite{yang2024qwen2} & 7B  & Instruct & \textcolor{green}{\ding{51}}& \textcolor{red}{\ding{55}}\\
        Sunsimiao-Qwen2-7B~\cite{Sunsimiao} & 7B &  - & \textcolor{green}{\ding{51}}& \textcolor{green}{\ding{51}}\\
        Yi-1.5-6B~\cite{young2024yi} & 6B  & Chat & \textcolor{green}{\ding{51}}& \textcolor{red}{\ding{55}}\\
        Yi-1.5-9B~\cite{young2024yi} & 9B  & Chat & \textcolor{green}{\ding{51}}& \textcolor{green}{\ding{55}}\\
        deepseek-llm-7b~\cite{deepseek-llm} & 7B  & chat & \textcolor{green}{\ding{51}}& \textcolor{red}{\ding{55}}\\
        gemma-2-9b~\cite{team2024gemma} & 9B & it & \textcolor{red}{\ding{55}}& \textcolor{red}{\ding{55}}\\
        granite-3.0-8b~\cite{granite2024granite} & 8B & instruct & \textcolor{red}{\ding{55}}& \textcolor{red}{\ding{55}}\\
        glm-4-9b~\cite{glm2024chatglm} & 9B & chat & \textcolor{green}{\ding{51}}& \textcolor{red}{\ding{55}}\\
        internlm2-7b~\cite{cai2024internlm2} & 7B & chat-7b & \textcolor{green}{\ding{51}}& \textcolor{red}{\ding{55}}\\
        internlm2.5-7b~\cite{cai2024internlm2} & 7B & chat-1m & \textcolor{green}{\ding{51}}& \textcolor{red}{\ding{55}}\\
        internlm3-8b~\cite{cai2024internlm2} & 8B & instruct & \textcolor{green}{\ding{51}}& \textcolor{red}{\ding{55}}\\
        \midrule
        \multicolumn{5}{c}{Open-source LLMs (server deployment)} \\
        \midrule
        Qwen2.5-72B~\cite{yang2024qwen2} & 72B & 72B-Instruct & \textcolor{green}{\ding{51}}& \textcolor{red}{\ding{55}}\\
        deepseek-v2.5-236B~\cite{liu2024deepseek} & 236B & - & \textcolor{green}{\ding{51}}& \textcolor{red}{\ding{55}}\\
        hunyuan-large-389B~\cite{sun2024hunyuan} & 389B & 2024-11-20 & \textcolor{green}{\ding{51}}& \textcolor{green}{\ding{55}}\\ 
        Meta-Llama-3.1-405B~\cite{dubey2024llama} & 405B & 405B-Instruct & \textcolor{red}{\ding{55}}& \textcolor{green}{\ding{55}}\\        
        deepseek-v3-671B~\cite{liu2025deepseek} & 671B & - & \textcolor{green}{\ding{51}}\\
        \midrule
        \multicolumn{5}{c}{Commercial LLMs} \\
        \midrule
        Spark4.0 Ultra & N/A & - & \textcolor{green}{\ding{51}}& \textcolor{red}{\ding{55}}\\
        claude-3-5-sonnet & 1750B & 20240620 & \textcolor{red}{\ding{55}}& \textcolor{red}{\ding{55}}\\
        doubao-pro & N/A & 128k-240628 & \textcolor{green}{\ding{51}}& \textcolor{green}{\ding{55}}\\
        ernie-4.0 & N/A & 8k-latest & \textcolor{green}{\ding{51}}& \textcolor{green}{\ding{55}}\\
        gemini-1.5-pro & N/A & 002 & \textcolor{red}{\ding{55}}& \textcolor{red}{\ding{55}}\\
        gemini-2.0-flash & N/A & exp & \textcolor{red}{\ding{55}}& \textcolor{green}{\ding{55}}\\
        glm-4-plus & N/A & - & \textcolor{green}{\ding{51}}& \textcolor{green}{\ding{55}}\\
        gpt-35-turbo & N/A & 0301 & \textcolor{red}{\ding{55}}& \textcolor{red}{\ding{55}}\\
        gpt-4o & N/A & 2024-08-06 & \textcolor{red}{\ding{55}}& \textcolor{red}{\ding{55}}\\
        grok-beta & 314B & - & \textcolor{red}{\ding{55}}& \textcolor{red}{\ding{55}}\\
        minimax & 6500B & abab6.5t-8k & \textcolor{green}{\ding{51}}& \textcolor{green}{\ding{55}}\\
        moonshot & N/A & v1-128k & \textcolor{green}{\ding{51}}& \textcolor{green}{\ding{55}}\\
        step-2 & N/A & 16k & \textcolor{green}{\ding{51}}& \textcolor{green}{\ding{55}}\\
        yi-large & N/A & - & \textcolor{green}{\ding{51}}& \textcolor{green}{\ding{55}}\\ 
        \bottomrule
    \end{tabular}
    \label{tab:model_description}
\end{table}

\begin{table*}[htb]
    \centering 
    \caption{OphthBench Leaderboard: Performance Comparison of $39$ LLMs using \textbf{Common} Prompts. The top $8$ values and the bottom $8$ values in each column are highlighted with red and green backgrounds, respectively.}
    \begin{threeparttable}
    \begin{tabular}{cl|c|ccccc|c}
    \toprule
    \textbf{Rank} & \textbf{Model} & \textbf{MMLU} & \textbf{Education} & \textbf{Triage} & \textbf{Diagnosis} & \textbf{Treatment} & \textbf{Prognosis} & \textbf{All scenarios} \\
    \midrule
    \textbf{1} & \textbf{hunyuan-large-389B} & \cellcolor{red!20}88.40\tnote{a} & \cellcolor{red!20}87.42 & \cellcolor{red!20}82.78 & \cellcolor{red!20}87.24 & \cellcolor{red!20}89.98 & \cellcolor{red!20}88.50 & 87.12 \\
    \textbf{2} & \textbf{deepseek-v3-671B} & \cellcolor{red!20}88.50\tnote{a} & \cellcolor{red!20}86.47 & \cellcolor{red!20}88.69 & \cellcolor{red!20}83.12 & \cellcolor{red!20}86.08 & 79.50 & 84.65 \\
    \textbf{3} & \textbf{gemini-2.0-flash} & - & 76.47 & \cellcolor{red!20}84.37 & \cellcolor{red!20}82.53 & \cellcolor{red!20}82.11 & \cellcolor{red!20}94.50 & 83.60 \\
    \textbf{4} & \textbf{Qwen2.5-72B} & \cellcolor{red!20}86.10\tnote{a} & \cellcolor{red!20}79.58 & \cellcolor{red!20}90.00 & 81.93 & \cellcolor{red!20}82.83 & \cellcolor{red!20}84.00 & 83.53 \\
    \textbf{5} & \textbf{gpt-4o} & \cellcolor{red!20}88.70\tnote{a} & \cellcolor{red!20}77.29 & \cellcolor{red!20}87.74 & 79.97 & \cellcolor{red!20}82.62 & \cellcolor{red!20}84.00 & 82.17 \\
    \textbf{6} & \textbf{glm-4-plus} & \cellcolor{red!20}86.80\tnote{a} & 75.34 & 81.89 & \cellcolor{red!20}85.53 & \cellcolor{red!20}84.82 & \cellcolor{red!20}84.00 & 82.14 \\
    \textbf{7} & \textbf{claude-3-5-sonnet} & \cellcolor{red!20}88.70\tnote{a} & 76.61 & \cellcolor{red!20}86.61 & \cellcolor{red!20}83.92 & 78.84 & \cellcolor{red!20}85.50 & 82.11 \\
    \textbf{8} & \textbf{doubao-pro} & - & \cellcolor{red!20}84.47 & 75.52 & \cellcolor{red!20}82.48 & \cellcolor{red!20}84.66 & \cellcolor{red!20}84.00 & 82.08 \\
    \textbf{9} & \textbf{step-2} & - & \cellcolor{red!20}86.31 & \cellcolor{red!20}85.55 & \cellcolor{red!20}83.11 & 64.76 & \cellcolor{red!20}88.50 & 80.58 \\
    \textbf{10} & \textbf{grok-beta} & \cellcolor{red!20}87.50\tnote{a} & 75.98 & 81.79 & 81.26 & 76.66 & 81.00 & 79.26 \\
    \textbf{11} & \textbf{ernie-4.0} & - & \cellcolor{red!20}76.65 & 73.47 & \cellcolor{red!20}83.19 & 76.76 & \cellcolor{red!20}85.50 & 78.86 \\
    \textbf{12} & \textbf{yi-large} & 83.80\tnote{a} & 69.59 & 82.44 & 79.99 & \cellcolor{red!20}80.17 & 79.50 & 78.06 \\
    \textbf{13} & \textbf{Meta-Llama-3.1-405B} & \cellcolor{red!20}88.60\tnote{a} & \cellcolor{red!20}80.85 & 79.03 & 68.03 & 75.70 & \cellcolor{red!20}87.00 & 77.61 \\
    \textbf{14} & \textbf{internlm3-8b} & 70.90\tnote{b} & 69.91 & 81.15 & 74.85 & 77.84 & \cellcolor{green!20}72.00 & 76.77 \\
    \textbf{15} & \textbf{minimax} & 78.70\tnote{a} & 68.60 & 79.71 & 78.04 & 74.40 & \cellcolor{red!20}84.00 & 76.59 \\
    \textbf{16} & \textbf{gemini-1.5-pro} & 85.90\tnote{a} & 72.23 & 72.61 & 78.50 & 78.09 & 81.00 & 76.33 \\
    \textbf{17} & \textbf{moonshot} & - & 68.84& 77.67 & 78.81 & 75.48 & 81.00 & 76.12 \\
    \textbf{18} & \textbf{Qwen2.5-7B} & 74.10\tnote{b} & 69.24 & \cellcolor{red!20}85.08 & 71.22 & 76.46 & 75.00 & 75.02 \\
    \textbf{19} & \textbf{Spark4.0 Ultra} & 85.80\tnote{a} & 72.94 & 78.57 & 76.03 & 67.53 & 78.00 & 74.38 \\
    \textbf{20} & \textbf{deepseek-v2.5-236B} & 80.40\tnote{a} & 65.36 & 78.84 & 77.09 & 68.51 & \cellcolor{green!20}72.00 & 72.00 \\
    \textbf{21} & \textbf{Yi-1.5-9B} & 69.70\tnote{b} & 68.68 & 73.15 & 71.22 & 71.12 & 75.00 & 71.77 \\
    \textbf{22} & \textbf{HuatuoGPT-o1-7B} & 65.20\tnote{b} & 67.37 & 80.49 & 71.02 & 72.97 & \cellcolor{green!20}66.00 & 71.22 \\
    \textbf{23} & \textbf{glm-4-9b} & 68.90\tnote{b} & 60.75 & 68.89 & 69.11 & 72.84 & 75.00 & 68.96 \\
    \textbf{24} & \textbf{internlm2\_5-7b} & 65.20\tnote{b} & 57.42 & 81.83 & 66.92 & 65.35 & 78.00 & 68.78 \\
    \textbf{25} & \textbf{Yi-1.5-6B} & \cellcolor{green!20}62.50\tnote{b} & 64.96 & 62.75 & 71.26 & 64.26 & 76.50 & 67.57 \\
    \textbf{26} & \textbf{Sunsimiao-Qwen2-7B} & 69.10\tnote{b} & 67.50 & 59.81 & \cellcolor{green!20}62.04 & 68.66 & 76.50 & 66.41 \\
    \textbf{27} & \textbf{gemma-2-9b} & 72.40\tnote{b} & 59.20 & 61.57 & 65.38 & 58.30 & \cellcolor{red!20}84.00 & 64.53 \\
    \textbf{28} & \textbf{internlm2-7B} & \cellcolor{green!20}62.40\tnote{b} & 49.25 & 72.97 & 66.33 & 58.04 & 75.00 & 62.78 \\
    \textbf{29} & \textbf{gpt-35-turbo} & 70.00\tnote{a} & 52.22 & \cellcolor{green!20}56.12 & 62.92 & 53.18 & 75.00 & 58.83 \\
    \textbf{30} & \textbf{Phi-3.5-mini} & 68.90\tnote{b} & 47.77 & 60.95 & \cellcolor{green!20}58.24 & \cellcolor{green!20}44.55 & 75.00 & 55.38 \\
    \textbf{31} & \textbf{Llama-3.1-Tulu-3-8B} & 65.10\tnote{b} & \cellcolor{green!20}35.97 & 57.42 & \cellcolor{green!20}59.12 & 52.30 & \cellcolor{green!20}66.00 & 51.86 \\
    \textbf{32} & \textbf{deepseek-llm-7b} & \cellcolor{green!20}51.50\tnote{b} & \cellcolor{green!20}40.52 & \cellcolor{green!20}50.64 & 64.86 & \cellcolor{green!20}45.11 & \cellcolor{green!20}63.00 & 51.08 \\
    \textbf{33} & \textbf{Meta-Llama-3.1-8B} & 67.80\tnote{b} & \cellcolor{green!20}40.88 & \cellcolor{green!20}50.08 & \cellcolor{green!20}55.76 & 4\cellcolor{green!20}6.42 & \cellcolor{green!20}67.50 & 50.65 \\
    \textbf{34} & \textbf{Mistral-7B} & 61.80\tnote{b} & 41.39 & \cellcolor{green!20}55.11 & \cellcolor{green!20}53.21 & \cellcolor{green!20}40.12 & 78.00 & 50.58 \\
    \textbf{35} & \textbf{Baichuan2-7B} & \cellcolor{green!20}52.90\tnote{b} & \cellcolor{green!20}38.34 & 64.46 & 64.62 & \cellcolor{green!20}47.03 & \cellcolor{green!20}46.50 & 50.08 \\
    \textbf{36} & \textbf{Ministral-8B} & \cellcolor{green!20}63.30\tnote{b} & \cellcolor{green!20}32.36 & \cellcolor{green!20}48.44 & \cellcolor{green!20}60.35 & 47.93 & \cellcolor{green!20}70.50 & 48.46 \\
    \textbf{37} & \textbf{granite-3.0-8b} & 64.60\tnote{b} & \cellcolor{green!20}33.16 & \cellcolor{green!20}56.81 & \cellcolor{green!20}53.73 & \cellcolor{green!20}36.64 & \cellcolor{green!20}43.50 & 42.86 \\
    \textbf{38} & \textbf{PULSE-7B} & \cellcolor{green!20}39.60\tnote{b} & \cellcolor{green!20}37.06 & 3\cellcolor{green!20}3.47 & \cellcolor{green!20}49.68 & \cellcolor{green!20}38.57 & \cellcolor{green!20}43.50 & 39.71 \\
    \textbf{39} & \textbf{HuatuoGPT2-7B} & \cellcolor{green!20}46.80\tnote{b} & \cellcolor{green!20}40.63 & \cellcolor{green!20}16.86 & 64.38 & \cellcolor{green!20}40.62 & \cellcolor{green!20}52.50 & 34.93 \\
    \bottomrule
    \end{tabular}

    \begin{tablenotes}
    \item[a] This value was retrieved from the Internet.
    \item[b] This value was collected through local experimental testing.
    \end{tablenotes}
    \end{threeparttable}
    \label{tab:common_performance}
\end{table*}

\begin{table*}[htb]
    \centering 
    \caption{OphthBench Leaderboard: Performance Comparison of $39$ LLMs using \textbf{Advanced} Prompts. The top $8$ values and the bottom $8$ values in each column are highlighted with red and green backgrounds, respectively.}
    \begin{threeparttable}
    \begin{tabular}{cl|c|ccccc|c}
    \toprule
    \textbf{Rank} & \textbf{Model} & \textbf{MMLU} & \textbf{Education} & \textbf{Triage} & \textbf{Diagnosis} & \textbf{Treatment} & \textbf{Prognosis} & \textbf{All scenarios} \\
    \midrule
    \textbf{1} & \textbf{deepseek-v3-671B} & \cellcolor{red!20}88.50\tnote{a} & \cellcolor{red!20}87.20 & \cellcolor{red!20}88.80 & \cellcolor{red!20}87.55 & \cellcolor{red!20}88.20 & \cellcolor{red!20}87.00 & 87.75 \\
    \textbf{2} & \textbf{step-2} & - & \cellcolor{red!20}84.71 & \cellcolor{red!20}86.24 & \cellcolor{red!20}86.75 & \cellcolor{red!20}82.82 & \cellcolor{red!20}88.50 & 85.76 \\
    \textbf{3} & \textbf{hunyuan-large-389B} & \cellcolor{red!20}88.40\tnote{a} & \cellcolor{red!20}85.78 & 82.89 & \cellcolor{red!20}86.47 & \cellcolor{red!20}87.65 & 82.50 & 85.01 \\
    \textbf{4} & \textbf{claude-3-5-sonnet} & \cellcolor{red!20}88.70\tnote{a} & \cellcolor{red!20}77.92 & \cellcolor{red!20}88.69 & \cellcolor{red!20}87.03 & \cellcolor{red!20}81.52 & 84.00 & 83.65 \\
    \textbf{5} & \textbf{glm-4-plus} & \cellcolor{red!20}86.80\tnote{a} & 76.52 & 80.08 & \cellcolor{red!20}88.92 & \cellcolor{red!20}83.87 & \cellcolor{red!20}88.50 & 83.30 \\
    \textbf{6} & \textbf{Qwen2.5-72B} & \cellcolor{red!20}86.10\tnote{a} & \cellcolor{red!20}80.35 & \cellcolor{red!20}85.74 & 83.85 & \cellcolor{red!20}81.13 & \cellcolor{red!20}84.00 & 82.97 \\
    \textbf{7} & \textbf{doubao-pro} & - & \cellcolor{red!20}82.87 & 81.23 & 83.12 & \cellcolor{red!20}83.93 & 82.50 & 82.72 \\
    \textbf{8} & \textbf{gpt-4o} & \cellcolor{red!20}88.70\tnote{a} & 75.55 & \cellcolor{red!20}85.94 & \cellcolor{red!20}84.79 & \cellcolor{red!20}83.53 & 82.50 & 82.29 \\
    \textbf{9} & \textbf{gemini-2.0-flash} & - & \cellcolor{red!20}77.68 & \cellcolor{red!20}84.84 & \cellcolor{red!20}86.09 & 80.66 & 82.50 & 82.25 \\
    \textbf{10} & \textbf{Meta-Llama-3.1-405B} & \cellcolor{red!20}88.60\tnote{a} & \cellcolor{red!20}78.56 & 80.49 & 79.81 & 79.44 & \cellcolor{red!20}90.00 & 81.46 \\
    \textbf{11} & \textbf{yi-large} & 83.80\tnote{a} & 75.01 & \cellcolor{red!20}84.84 & 83.23 & 80.49 & 79.50 & 80.47 \\
    \textbf{12} & \textbf{ernie-4.0} & - & 75.61\tnote{a} & 80.03 & 83.57 & 76.00 & \cellcolor{red!20}85.50 & 79.95 \\
    \textbf{13} & \textbf{Spark4.0 Ultra} & 85.80\tnote{a} & 72.95 & 80.60 & 81.89 & 80.65 & \cellcolor{red!20}84.00 & 79.83 \\
    \textbf{14} & \textbf{grok-beta} & \cellcolor{red!20}87.50\tnote{a} & 74.76 & 80.14 & \cellcolor{red!20}85.51 & 75.94 & 82.50 & 79.57 \\
    \textbf{15} & \textbf{deepseek-v2.5-236B} & 80.40\tnote{a} & 72.55 & 83.77 & 84.26 & 72.30 & \cellcolor{red!20}85.50 & 79.22 \\
    \textbf{16} & \textbf{internlm3-8b} & 70.90\tnote{b} & 69.91 & 81.15 & 74.85 & 79.20 & \cellcolor{green!20}72.00 & 76.77 \\
    \textbf{17} & \textbf{gemini-1.5-pro} & 85.90\tnote{a} & 69.04 & 78.63 & 78.90 & 74.44 & \cellcolor{red!20}84.00 & 76.67 \\
    \textbf{18} & \textbf{minimax} & 78.70\tnote{a} & 66.70 & 82.44 & 82.47 & 72.22 & 79.50 & 76.13 \\
    \textbf{19} & \textbf{Qwen2.5-7B} & 74.10\tnote{b} & 64.68 & 81.65 & 73.35 & 80.05 & 82.50 & 75.81 \\
    \textbf{20} & \textbf{Yi-1.5-9B} & 69.70\tnote{b} & 70.01 & 71.40 & 73.51 & 69.09 & \cellcolor{red!20}88.50 & 73.89 \\
    \textbf{21} & \textbf{internlm2.5-7b} & 65.20\tnote{b} & 57.40 & 76.77 & 69.28 & 71.36 & 82.50 & 70.41 \\
    \textbf{22} & \textbf{moonshot} & - & 69.66& \cellcolor{red!20}87.82 & 74.39 & \cellcolor{green!20}49.81 & 82.50 & 70.05 \\
    \textbf{23} & \textbf{HuatuoGPT-o1-7B} & 65.20\tnote{b} & 64.55 & 84.83 & 67.20 & 64.62 & \cellcolor{green!20}69.00 & 69.32 \\
    \textbf{24} & \textbf{glm-4-9b} & 68.90\tnote{b} & 53.76 & 75.75 & 74.77 & 73.33 & 73.50 & 69.04 \\
    \textbf{25} & \textbf{Sunsimiao-Qwen2-7B} & 69.10\tnote{b} & 62.88 & 68.00 & \cellcolor{green!20}63.74 & 66.24 & \cellcolor{red!20}84.00 & 68.21 \\
    \textbf{26} & \textbf{Yi-1.5-6B} & \cellcolor{green!20}62.50\tnote{b} & 62.80 & 67.57 & 69.46 & 60.18 & 79.50 & 67.28 \\
    \textbf{27} & \textbf{internlm2-7B} & \cellcolor{green!20}62.40\tnote{b} & 52.50 & 73.96 & 67.34 & 60.52 & 73.50 & 64.48 \\
    \textbf{28} & \textbf{gemma-2-9b} & 72.40\tnote{b} & 58.24 & 64.83 & 66.90 & 61.54 & \cellcolor{green!20}72.00 & 64.37 \\
    \textbf{29} & \textbf{gpt-35-turbo} & 70.00\tnote{a} & 51.25 & 74.59 & \cellcolor{green!20}65.11 & 56.58 & 79.50 & 63.67 \\
    \textbf{30} & \textbf{Llama-3.1-Tulu-3-8B} & 65.10\tnote{b} & 50.07 & 65.53 & 66.84 & 53.54 & 81.00 & 61.56 \\
    \textbf{31} & \textbf{Meta-Llama-3.1-8B} & 67.80\tnote{b} & 52.44 & \cellcolor{green!20}61.35 & \cellcolor{green!20}63.79 & \cellcolor{green!20}49.61 & \cellcolor{green!20}69.00 & 58.35 \\
    \textbf{32} & \textbf{Phi-3.5-mini} & 68.90\tnote{b} & \cellcolor{green!20}47.62 & 66.85 & \cellcolor{green!20}62.20 & \cellcolor{green!20}46.90 & 79.50 & 58.18 \\
    \textbf{33} & \textbf{deepseek-llm-7b} & \cellcolor{green!20}51.50\tnote{b} & \cellcolor{green!20}42.78 & \cellcolor{green!20}62.67 & 65.37 & \cellcolor{green!20}47.72 & 76.50 & 56.40 \\
    \textbf{34} & \textbf{Ministral-8B} & \cellcolor{green!20}63.30\tnote{b} & \cellcolor{green!20}38.53 & \cellcolor{green!20}56.83 & 67.14 & 52.30 & 76.50 & 55.17 \\
    \textbf{35} & \textbf{Baichuan2-7B} & \cellcolor{green!20}52.90\tnote{b} & \cellcolor{green!20}45.92 & \cellcolor{green!20}51.06 & 65.26 & 55.80 & \cellcolor{green!20}48.00 & 52.39 \\
    \textbf{36} & \textbf{Mistral-7B} & \cellcolor{green!20}61.80\tnote{b} & \cellcolor{green!20}47.16 & \cellcolor{green!20}45.86 & \cellcolor{green!20}54.34 & \cellcolor{green!20}43.45 & \cellcolor{green!20}67.50 & 50.38 \\
    \textbf{37} & \textbf{PULSE-7B} & \cellcolor{green!20}39.60\tnote{b} & \cellcolor{green!20}32.67 & \cellcolor{green!20}49.95 & \cellcolor{green!20}63.17 & \cellcolor{green!20}41.53 & \cellcolor{green!20}66.00 & 47.31 \\
    \textbf{38} & \textbf{HuatuoGPT2-7B} & \cellcolor{green!20}46.80\tnote{b} & \cellcolor{green!20}41.78 & \cellcolor{green!20}35.69 & \cellcolor{green!20}63.57 & \cellcolor{green!20}49.34 & \cellcolor{green!20}49.50 & 46.23 \\
    \textbf{39} & \textbf{granite-3.0-8b} & 64.60\tnote{b} & \cellcolor{green!20}34.13 & \cellcolor{green!20}50.94 & \cellcolor{green!20}60.93 & \cellcolor{green!20}47.43 & \cellcolor{green!20}39.00 & 44.62 \\
    \bottomrule
    \end{tabular}

    \begin{tablenotes}
    \item[a] This value was retrieved from the Internet.
    \item[b] This value was collected through our experimental testing.
    \end{tablenotes}
    \end{threeparttable}
    \label{tab:advanced_performance}
\end{table*}

Table~\ref{tab:common_performance} and \ref{tab:advanced_performance} present ranking list for common and advanced performance respectively.
With a comprehensive analysis, we draw out the following findings.

\begin{figure}[!tb]
    \centering
    \includegraphics[width=\linewidth]{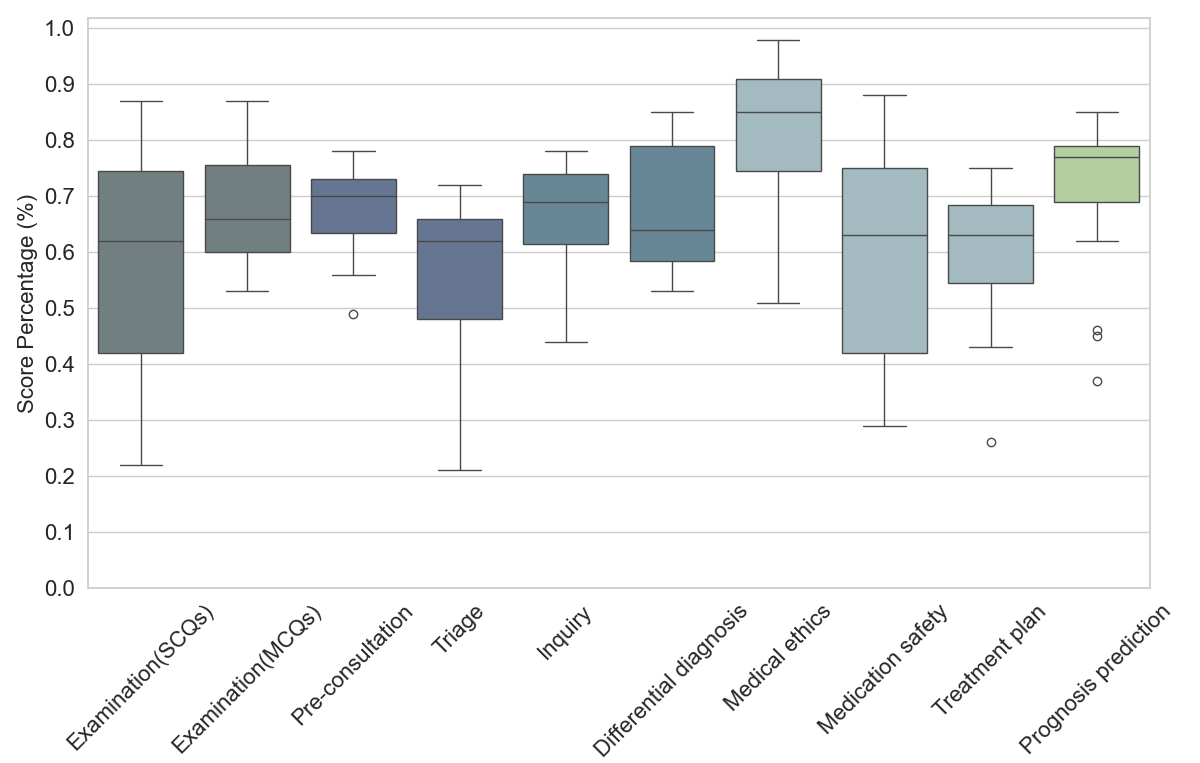}
    \caption{\textbf{A statistic of scoring rate of $39$ for each task.} The results are collected with an advanced prompt to illustrate the practice challenge better.}
\label{fig:gap}
\end{figure}

\textbf{LLMs currently face significant challenges in practical applications within ophthalmology.} 
As shown in Fig.~\ref{fig:gap}, both open-source and commercial large language models (LLMs) demonstrate a performance rate of approximately $70\%$, underlining the difficulties these models face when applied to ophthalmic tasks. 
Among the five scenarios assessed, the prognosis scenario stands out as the most promising, while the education scenario remains notably more challenging, reflecting the varied levels of performance across different ophthalmic applications.
Fortunately, as data quality and training techniques continue to evolve, their potential utility in ophthalmology shows an encouraging improvement. 
This is evident from the progress observed in the \texttt{internlm} series~\cite{cai2024internlm2}, where the model ranking improved from $28$th (\texttt{internlm2}) to $14$th (\texttt{internlm3}) in Table~\ref{tab:common_performance} and from $27$th (\texttt{internlm2}) to $16$th (\texttt{internlm3}) in Table~\ref{tab:advanced_performance}.
These advancements 

\textbf{LLMs developed by Chinese companies or institutions demonstrate superior performance in OphthBench.} 
Although LLMs tested in our experiments could handle multiple languages, those developed by Chinese companies or institutions consistently outperformed others. 
As shown in Table~\ref{tab:common_performance} and Table~\ref{tab:advanced_performance}, the majority of top-performing models were developed by Chinese companies or institutions. 
Further analysis of models with approximately $7$ billion parameters ($6$B to $9$B) shows a performance gap, with non-Chinese models scoring lower. 
While various factors, such as model selection, question design, and prompt formulation, may influence the results, the t-test reveals a marginally significant difference with $P = 0.096$ for Table~\ref{tab:common_performance} and $P = 0.085$ for Table~\ref{tab:advanced_performance}.
We hypothesize that the poorer performance of non-Chinese models can be attributed to the following factors:
Firstly, the relatively smaller proportion of Chinese data used in their training may limit their effectiveness. 
More importantly, we believe that the distinct differences in medical systems between China and other regions could hinder the models' ability to generalize across diverse ophthalmic contexts, ultimately impacting their performance.

\textbf{Medical LLMs do not demonstrate superior ophthalmic capabilities when compared to general LLMs.} 
As shown in Table~\ref{tab:common_performance} and Table~\ref{tab:advanced_performance}, PULSE-7B, Sunsimiao-Qwen2-7B, and HuatuoGPT2-7B are below than $61.5$\% LLMs. 
HuatuoGPT2-7B which was developed based on Baichuan2-7B and HuatuoGPT2-o1-7B which developed based on Qwen2.5-7B both do not bet down their foundation model.
We believe that it is attributed to two facets: one is their development technique is outdated, where PULSE-7B, Sunsimiao-Qwen2-7B, and HuatuoGPT2-7B were developed in 2023.
Another is the specialized knowledge needed in Ophthalmology does not or is little covered in their training data.
Hence, the learned knowledge is forgotten (namely degeneration) during their continued post-training.

\begin{figure*}[!tb]
    \centering
    \includegraphics[width=\linewidth]{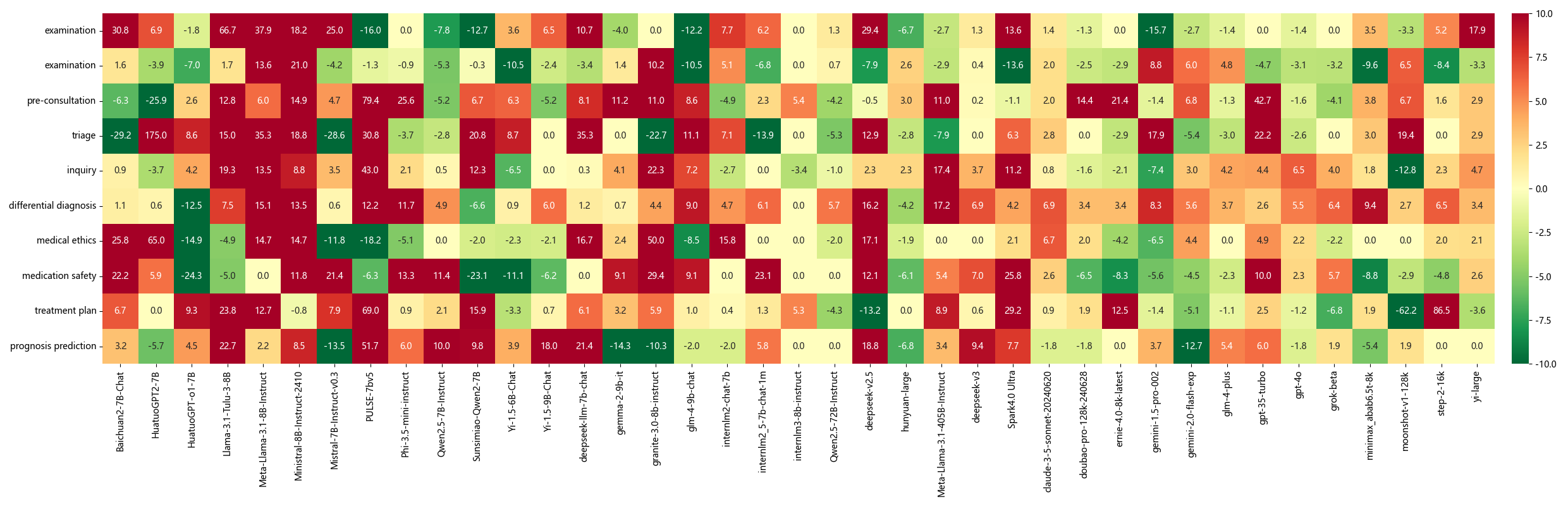}
    \caption{\textbf{Impact of Prompts.} We compared the responses generated by common and advanced prompts, reporting the performance enhancement across various tasks and models. Notably, the advanced prompts resulted in apparent performance improvements ($\geq 10\%$) in most cases, with the highest increase reaching up to $175.0\%$. However, since the advanced prompts utilized in this paper were not tailored to specific models or tasks, a performance decline of $62.2\%$ was also observed.}
\label{fig:confusion}
\end{figure*}

\textbf{Prompt engineering can optimize the utilization of LLMs in ophthalmic practice.}
As shown in Fig.~\ref{fig:prompt}, we employed two types of prompts to generate common and advanced responses, respectively. 
The results presented in Table~\ref{tab:common_performance} and Table~\ref{tab:advanced_performance} demonstrate that the performance of most models improved, even though the prompts were not specifically tailored to each model or question.
A more detailed analysis is presented in Fig.~\ref{fig:confusion}, which demonstrates that the performance of LLMs is enhanced in most cases through the use of sophisticated prompting techniques.
Based on these findings, we contend that prompt engineering is essential for optimizing the application of LLMs in ophthalmology.

\section{Discussion}
To support the real-world application and advance the development of large language models in Chinese ophthalmology, we introduce a specialized benchmark, OphthBench, designed to assess LLM performance within the context of Chinese ophthalmic practices. 
OphthBench offers several key advantages over previous LLM benchmarking efforts:
(1) Practicality: Developed by three ophthalmologists, OphthBench is meticulously constructed to align with real-world ophthalmic practices, ensuring its accurate assessment in practicality evaluation.
(2) Diversity of question types: The benchmark includes three question types, single-choice questions, multiple-choice questions, and open-ended questions, as well as a dedicated scoring strategy, avoiding the one-dimensionality often seen in other benchmarks.
(3) Prompt design: OphthBench uses two distinct prompt versions, one for asking the LLMs to generate common responses to simulate the common users, and another one designed to aid the LLMs generate advanced responses to maximize their performance, to provide objective and multifaceted assessments.

After conducting extensive experiments and in-depth analysis of $39$ popular LLMs on this benchmark, we draw several conclusions regarding their real-world applications. 
Consistent with broader trends in LLM research, we observe that improvements in model performance are closely linked to increases in the number of parameters and general capabilities, as measured by benchmarks like MMLU. 
This supports the validity of the OphthBench benchmark (see Table~\ref{tab:common_performance} and \ref{tab:advanced_performance}).

However, this study also has limitations: it does not include all LLMs, especially ophthalmology-specific models, many of which are currently inaccessible. 
Additionally, the dataset used is limited in both size and the types of tasks it covers. 
Future work could extend the question set under this framework to address these limitations.

\section{Conclusion}
In this paper, we present OphthBench, a specialized benchmark designed to assess LLM capabilities in Chinese ophthalmic practices. 
It focuses on $5$ core ophthalmic scenarios, \textbf{Education}, \textbf{Triage}, \textbf{Diagnosis}, \textbf{Treatment}, and \textbf{Prognosis}, covering $9$ distinct tasks with single-choice, multiple-choice, and open-ended question formats.
Evaluation results from $39$ widely-used LLMs on OphthBench highlight a substantial gap between current model capabilities and practical requirements. 
Meanwhile, the effectiveness of OphthBench is confirmed through its discernible scoring for scenarios and models.
We hope that OphthBench will not only facilitate the standardization and refinement of LLM applications for Chinese ophthalmology but also inspire the development of more advanced and domain-specific LLMs in the future.

\bibliographystyle{unsrt}
\bibliography{references}  






\end{document}